# Understanding eGFR Trajectories and Kidney Function Decline via Large Multimodal Models

Chih-Yuan Li
*Department of Information Management*
*National Sun Yat-sen University*
Kaohsiung, Taiwan
stellali1801@gmail.com

Jun-Ting Wu
*Department of Information Management*
*National Sun Yat-sen University*
Kaohsiung, Taiwan
lydiazwu@gmail.com

Chan Hsu
*Department of Information Management*
*National Sun Yat-sen University*
Kaohsiung, Taiwan
chanshsu@gmail.com

Ming-Yen Lin
*Kaohsiung Medical University Hospital*
*Kaohsiung Medical University*
Kaohsiung, Taiwan
mingyenlin3@gmail.com

Yihuang Kang
*Department of Information Management*
*National Sun Yat-sen University*
Kaohsiung, Taiwan
ykang@mis.nsysu.edu.tw

*Abstract*—The estimated Glomerular Filtration Rate (eGFR) is an essential indicator of kidney function in clinical practice. Although traditional equations and Machine Learning (ML) models using clinical and laboratory data can estimate eGFR, accurately predicting future eGFR levels remains a significant challenge for nephrologists and ML researchers. Recent advances demonstrate that Large Language Models (LLMs) and Large Multimodal Models (LMMs) can serve as robust foundation models for diverse applications. This study investigates the potential of LMMs to predict future eGFR levels with a dataset consisting of laboratory and clinical values from 50 patients. By integrating various prompting techniques and ensembles of LMMs, our findings suggest that these models, when combined with precise prompts and visual representations of eGFR trajectories, offer predictive performance comparable to existing ML models. This research extends the application of foundation models and suggests avenues for future studies to harness these models in addressing complex medical forecasting challenges.

*Keywords—Large Multimodal Model, estimated Glomerular Filtration Rate, Healthcare, Chronic Kidney Disease*

## I. INTRODUCTION

Chronic Kidney Disease (CKD) poses a significant global health challenge, affecting more than 850 million people worldwide [1]. The estimated Glomerular Filtration Rate (eGFR) serves as a crucial measurement for assessing kidney function and monitoring CKD progression. Precise predictions of eGFR trajectories are critical, enabling timely intervention, tailored treatment plans, and improved patient outcomes [2], [3]. Understanding eGFR trajectories also facilitates healthcare providers to manage kidney function effectively and arrange time-critical interventions appropriately. For instance, accurately predicting the progression to end-stage renal disease (ESRD) enables patients and nephrologists to prepare patients' vascular access 3 to 6 months before initiating hemodialysis, highlighting the necessity for advanced predictive models.

Traditionally, eGFR is estimated using formulas like the Modification of Diet in Renal Disease (MDRD) Study equation [4] and the Chronic Kidney Disease Epidemiology Collaboration equation (CKD-EPI) [5], which incorporate serum creatinine levels and demographic factors to calculate eGFR [6]. While these equations form a robust clinical foundation and are widely utilized, they lack the capability to predict the future eGFR. Recent machine learning (ML) models have explored the dynamic selection of estimation equations [7], ML-based alternatives to estimate GFR [8], and predicting the risks of progression of CKD and the need for kidney transplantation [9], [10]. However, these models often do not address the prediction of future eGFR trajectories, which are influenced by both linear and nonlinear patient-specific patterns [11]. Traditional models, such as group-based trajectory and mixed-effect models, frequently fail to capture this complexity, underscoring the urgent need for more sophisticated predictive tools.

In response to these challenges, the emergence of Large Language Models (LLMs) and Large Multimodal Models (LMMs), such as GPT-4 [12] and Google Gemini [13], marks a significant advancement. Trained on extensive datasets, these models demonstrate exceptional capabilities as foundation models for a variety of downstream tasks without the need for fine-tuning [14]. The rich, implicit knowledge embedded in LLMs and LMMs enables these models to enhance input data, offering a novel approach to enrich electronic health record information and provide more accurate eGFR trajectory predictions than conventional ML models.

This paper explores the capability of LMMs to precisely predict eGFR, thereby enabling clinicians to diagnose and manage CKD with enhanced accuracy. Leveraging their advanced capabilities in processing and generating human language across diverse domains, we hypothesize that LMMs can outperform traditional predictive models like Random Forest [15] and 1D-CNN [16]. Through innovative techniques such as prompt engineering, ensembles of LMMs, and repeated questioning, we aim to refine the accuracy and clinical utility of eGFR predictions. The key contributions of this study are summarized as follows:

- Demonstrating the potential of LMMs to improve the accuracy of eGFR predictions.

- Utilizing prompt engineering to optimize the effectiveness of LMMs in eGFR forecasting.

- Evaluating the performance of LMM ensembles in enhancing eGFR prediction accuracy.

The remainder of this paper is organized as follows: Section 2 reviews existing prediction methods for eGFR, emphasizing the advancements and limitations of both traditional and computational approaches. Section 3 details our proposed framework. Section 4 presents the results of our experiments, comparing the performance across LLMs and conventional ML models. Section 5 discusses the clinical implications of our findings. Finally, Section 6 concludes with future research directions in the computational models of eGFR prediction.

## II. Background

The eGFR is one of the most frequently used metrics to represent patient kidney function in clinical practice. Serving as a key indicator of renal performance, eGFR is derived from serum creatinine levels alongside demographic variables such as age, sex, and race [5]. This estimation method stands out due to its non-invasive nature and ease of calculation, offering a pragmatic alternative to direct measurement techniques, which are often invasive, expensive, and time-consuming.

A decrease in eGFR is one important indicator of kidney function impairment, signaling a reduced ability of the kidneys to filter out waste and excess fluid from the blood. This decline is not merely a flag for potential renal pathology but also serves as a prognostic metric for risks associated with End-Stage Kidney Disease (ESKD), cardiovascular incidents, and increased mortality.

eGFR's central role in nephrology is instrumental in diagnosing and monitoring kidney health [17], [18]. It is the primary tool used to assess the existence and severity of CKD [3], to keep track of disease progression, and to tailor medication dosages and medical procedures such as renal replacement therapy to individual patient needs. The staging of CKD relies heavily on eGFR, with each stage corresponding to a specific eGFR range, thereby guiding the clinical decision to treatment and management.

The significance of understanding eGFR trajectories extends into the realm of preemptive healthcare [19]. Accurate eGFR modeling enables early interventions that can effectively slow CKD progression, avert relevant complications, and prepare for renal replacement therapy. The predictive value of eGFR is thus pivotal—not only does it assist in diagnosing renal issues, but it also helps in crafting a patient's treatment journey over time, thereby improving the quality of care delivered.

The dynamic nature of kidney function necessitates a model that transcends static readings. A longitudinal perspective afforded by eGFR trajectories offers a richer, more contextual understanding of renal function over time. Hence, leveraging large multimodal datasets through sophisticated computational techniques is an emerging frontier in the nuanced exploration of eGFR patterns.

In summary, eGFR is not just a critical indicator of current renal function but also triggers consequent clinical diagnosis, treatment, and care. It is instrumental in translating complex patient data into actionable clinical insights. Advancing the accuracy of eGFR predictions with robust multimodal models holds substantial promise for improving patient outcomes and assisting clinicians in making correct clinical decisions.

The traditional methodologies employed for estimating eGFR have historically relied on serum creatinine as a staple biomarker. This has led to the derivation of seminal formulas like the Cockcroft-Gault and the Modification of Diet in Renal Disease (MDRD) Study equation. By integrating creatinine readings with demographic variables such as age, sex, and body size, these equations form the bedrock of eGFR estimation. Nonetheless, they are not without their drawbacks. The accuracy of these formulas can be compromised by patient-specific factors such as muscle mass, dietary intake, and co-existing health conditions, necessitating the development of more nuanced approaches.

The advent of machine learning models has marked a transformative period in eGFR prediction. Advanced algorithms have expanded the horizons of variable integration, potentially enhancing the precision of renal function assessments [10]. Existing studies have widely employed ML models to predict GFR and CKD stages. Wang et al. developed a two-branches neural network to predict both log GFR and CKD stages simultaneously [20]. By optimizing for the two targets, they achieved competitive performance in estimating GFR and classifying CKD stages compared to traditional equations. Fan et al. proposed to select optimistic equations to estimate different patients' GFR with decision trees [21]. Their results show that the proposed decision trees can reduce errors in estimated GFR compared to using one of the other 13 equations to estimate GFR. Furthermore, research by Inaguma et al. developed ML models to identify extremely rapid declines in eGFR [22]. They first partitioned patients into three groups by the starting point of eGFR and further partitioned each group into three subgroups by the trajectory of eGFR. Then, they built random forests to classify acute decline in eGFR for each group and achieve the AUCs range from 0.694 to 0.788.

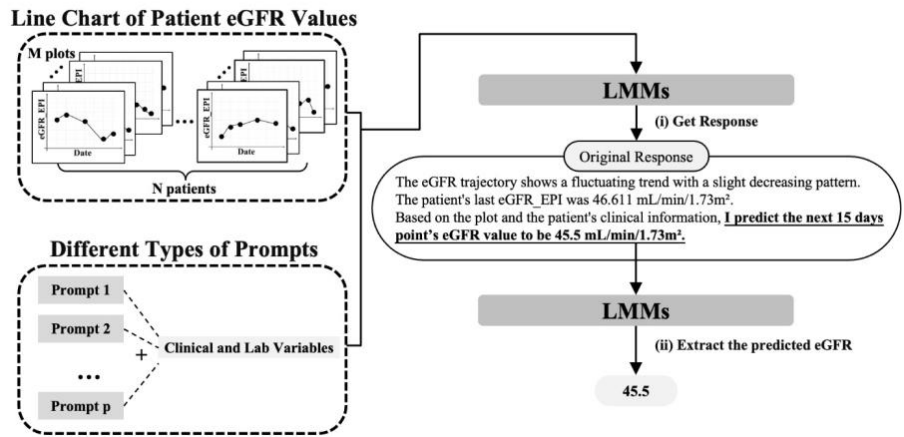

Fig. 1. Framework Overview: *M* plots represent the data and plots of each patient at different time points, where the number of data points is determined by the median of all patients' data. *N* represents the total number of patients, and p denotes the index of each.

Building upon the advancements of traditional ML models, the introduction of LMMs has further revolutionized healthcare, a trend that cannot be ignored [23], [24]. One of the most notable contributions of LMMs is their ability to assist healthcare professionals in the diagnostic process [19]. However, the integration of LMMs in healthcare is not without challenges [25]. Ensuring data privacy, managing the complexity of medical terminologies, and He et al. suggest that addressing potential biases in the training data are critical issues that need to be resolved to fully realize the potential of LMMs in clinical settings [23].

The integration of LMMs into healthcare holds immense promise for enhancing diagnostic processes, personalizing treatment regimens, and ultimately improving overall patient care. Leveraging their advanced capabilities in understanding and generating language, as well as analyzing images, these models stand poised to provide invaluable support to healthcare professionals. Moreover, LMMs have demonstrated diverse contributions within the healthcare domain, with their applications expanding rapidly. Nevertheless, their utilization in tabular data analysis remains relatively underexplored. Hence, tapping into the growing versatility of LMMs, we aim to leverage their capabilities to comprehend tabular data and predict eGFR, further expanding their utility beyond image-based analyses in healthcare contexts.

## III. METHODS

Our framework aims to utilize LMMs to enhance the predictive performance in forecasting eGFR for future patient visits. The framework systematically incorporates patient data, clinical and laboratory variables, and structured prompts through iterative processing. These repeated experiments ensure robustness and consistency in the predictions, providing a reliable tool for clinicians in managing patient care.

In the initial step, depicted in the left segment of Figure 1, we input N patients' line charts, clinical and laboratory variables, along with various prompts, into the LMMs to derive eGFR predictions. Each patient has $M$ plots, where $M$ is determined by the median count of data points available per patient. Each patient's m-th plot includes one more data point than the (m-1)-th plot, specifically an additional eGFR value and its corresponding date. We sequentially process each patient's plots to predict the next eGFR value in each plot. All the data encompassed within the $M$ plots will be utilized by both the Random Forest and 1D-CNNs; they can also access past data.

In this task, a series of prompts were meticulously designed to guide the LMMs in interpreting the data. For example, the prompts included context-specific scenarios, fill-in-the-blank formats, and other variations to capture diverse nuances in the data [29], [30], [31]. After obtaining responses from the LMMs, as illustrated in part (i) of the figure, the unstructured responses are reprocessed by the LMMs to extract the predicted values, as shown in part (ii) of the figure.

This process is repeated multiple times, each iteration combining patient data, clinical and laboratory variables, and new prompts. We process each patient's plots to ensure robustness and consistency in the predictions. To validate the predictions, we average the multiple predictions for the same plot and prompt, evaluating whether the overall results have improved. Through these iterative steps and the averaging of results, we aim to leverage the comprehensive insights provided by the LMMs to refine and enhance the predictive accuracy of our models. Additionally, we implemented prompt ensemble and LMM ensemble techniques. In these text-based responses, we further utilized LMMs and functions to extract the predicted eGFR values.

The performance of the LMMs was compared with traditional machine learning models such as Random Forest and 1D-CNNs, which were only provided with clinical and laboratory variable data for predictions. Statistical tests were conducted to assess the significance of performance improvements. The comparison focused on the accuracy of eGFR predictions and the models' ability to capture the intricate dynamics of kidney function.

## IV. EXPERIMENTS

We selected study data through using the Kaohsiung Medical University Hospital Research Databases, 2004-2021.

TABLE I. VARIABLES IN DATASET

| Variables Category | Variables |
|---|---|
| Demographic | gender, age, cause of Chronic Kidney Disease, smoking, drinking frequency, CKD stage according to eGFR_EPI. |
| Laboratory measurements | eGFR based on the CKD-EPI equation (eGFR), blood urea nitrogen (BUN), phosphorus, and urine albumin-to-creatinine ratio (UACR). |
| Comorbidity | Diabetes Mellitus, Hypertension, Hyperlipidemia, Coronary Artery Disease, Cardiovascular Disease, Atrial Fibrillation, Peripheral Artery Disease, Dementia, Hepatitis B Virus, Hepatitis C Virus, Liver Cirrhosis, Peptic Ulcer, Malignancy, Gouty Nephropathy, Chronic Obstructive Pulmonary Disease, Asthma, and Charlson Comorbidity Index. |
| Medication | Antiplatelets, Anticoagulants, ACE Inhibitors or ARBs, Calcium Channel Blockers, Beta-Blockers, Alpha-Blockers, Statins, Fibrates, Metformin, DPP4 Inhibitors, Thiazolidinediones, Sulfonylureas, Alpha-GlucosidaseInhibitors, Insulin, Proton Pump Inhibitors, H2 Blockers, Thiazide Diuretics, Loop Diuretics, Potassium-Sparing Agents, Colchicine, Uric Acid Lowering Agents, NSAIDs, Traditional NSAIDs, and COX2 Inhibitors. |

Patients enrolled in the pre-ESRD care program at two Kaohsiung Medical University affiliated hospitals over three months were identified from the databases. The details of the care program have been published in our previous study [32]. Briefly, the patients who were involved in the care program are required to have regular eGFR values evaluation for reimbursement. We first identified the patients by the specific reimbursed codes from the databases. Then, we defined the index date as the care enrollment date and traced all the patients' laboratory records. Through the above process, we identified 10,755 patients with 402,970 records of eGFR measurements during the study period. Subsequently, we conducted data preprocessing, which included removing data recorded during patients' hospitalization periods. To ensure model quality, we further excluded records where the interval between visits exceeded one year and eliminated records with fewer than five data points. In the preliminary experiment, we randomly selected 50 patients with a total of 564 eGFR measurements that were further tested. The study was approved by the Institutional Review Board of Kaohsiung Medical University Hospital (KMUHIRB-EXEMPT(I)-20210123).

To explore the effectiveness of our framework in a wide range of patients with different initiated eGFR. All the patient's eGFR within databases are included. This dataset included various categories of variables, such as demographic information, laboratory measurements in each visit, common comorbidity, and the selected medications for chronic diseases; more detailed information could be attached in Table 1. Cause of CKD and life habitant information was obtained when patients enrolled in the care program. The frequency of the selected laboratory data was ordered by physicians based on patient eGFR value. In general, CKD should be checked eGFR per three months. We determined the study comorbidities by a series of diagnosis codes from outpatient medical records. Common medication prescriptions in CKD were identified by outpatient records. Once patients were prescribed the selected medication they were considered to use the medication. The baseline eGFR of these patients on their first visit ranges from 2.44 to 171.85 with a mean value of 43.31, and the patients are aged from 19.5 to 87.6 with a mean value of 63.1.

We initiated our inquiry with the LMMs, presenting them with line charts depicting the eGFR trajectory alongside prompts generated from a manually crafted template. Each line chart encapsulates the eGFR data gathered from multiple visits at distinct time intervals for individual patients, as illustrated in Figure 2. These visual representations are instrumental in enabling the LMMs to comprehend and predict patterns within the data. The x-axis of the plot delineates the dates corresponding to each eGFR measurement, while the y-axis quantifies the eGFR in mL/min/1.73m² units. This graphical representation offers a lucid depiction of a patient's kidney function over time, thereby enhancing the understanding of chronic kidney disease progression.

TABLE II. COMPARISON OF MODEL PERFORMANCE WITH DIFFERENT FEATURES.

| Model | Prompt | Train | | Validation | |
|---|---|---|---|---|---|
| | | *MAE* | *MAPE(%)* | *MAE* | *MAPE(%)* |
| RF | - | 1.20 | 4.25 | 2.76 | 12.26 |
| 1D-CNN | - | 2.88 | 14.16 | 5.49 | 15.70 |
| Gemini Flash | 1 | 10.79 | 24.89 | 12.65 | 30.34 |
| | **2** | **1.03** | **3.59** | **1.58** | **5.57** |
| | 3 | 8.20 | 42.88 | 15.44 | 78.59 |
| | 4 | 3.84 | 29.03 | 3.87 | 22.14 |
| | ensemble | 4.14 | 20.30 | 5.38 | 23.90 |
| Gemini Pro Vision | **1** | **1.20** | **5.47** | **1.54** | **4.98** |
| | 2 | 1.44 | 6.60 | 2.17 | 6.87 |
| | 3 | 5.26 | 22.17 | 7.23 | 30.43 |
| | 4 | 2.64 | 11.62 | 3.27 | 14.29 |
| | ensemble | 1.82 | 8.55 | 2.81 | 11.38 |
| GPT-4o | 1 | 2.04 | 7.79 | 2.06 | 7.31 |
| | 2 | 3.12 | 13.39 | 4.16 | 30.68 |
| | 3 | 5.93 | 39.81 | 4.89 | 24.43 |
| | 4 | 3.03 | 12.97 | 3.28 | 14.67 |
| | ensemble | 2.96 | 16.74 | 3.24 | 17.11 |
| Claude 3 Opus (self-moderated) | 1 | 2.87 | 12.56 | 3.85 | 15.08 |
| | 2 | 2.77 | 12.40 | 3.59 | 16.87 |
| | 3 | 2.89 | 11.64 | 3.70 | 16.65 |
| | 4 | 3.13 | 13.48 | 3.85 | 17.41 |
| | ensemble | 2.62 | 11.22 | 3.23 | 14.07 |

One of the prompts in Appendix is: **"Based on the provided plot, the x-axis represents the dates of each eGFR measurement, while the y-axis shows the eGFR values in units of mL/min/1.73m². This plot depicts the trajectory of a single patient's kidney function, as measured by the estimated glomerular filtration rate (eGFR). Please fill in the blank: The most likely predicted value for the next {next_day_diff} days is {{eGFR}}mL/min/1.73m². The latest data for the patient: {data_text}.".** Within the prompt, placeholders for **{next_day_diff}** and **{data_text}** are employed to personalize the inquiry according to the specific patient's data. The value within **{next_day_diff}** is computed from the last recorded data point to the time of eGFR prediction, indicating the expected eGFR in the upcoming days. This method amplifies the precision of questioning by integrating the most recent data available. **{data_text}** is replaced by variable names and their values in the 12th visit, including blood urea nitrogen (BUN), phosphorus, urine albumin-to-creatinine ratio (UACR), and eGFR in the latest 3 visits, along with other pertinent details. After presenting the LMMs with line charts and prompts, we further leveraged functions and LMMs to extract predicted eGFR from the textual responses generated by the LMMs, the remaining prompts will be included in the Appendix.

In our experiment, we adopt a train-test split strategy where 70% of the patients are randomly selected for training, while the remaining 30% are reserved for testing. This ensures that the models are evaluated on unseen data, providing a reliable measure of their generalizability and effectiveness.

After applying our framework and prompts to the dataset, we observed promising results in predicting eGFR for future patient visits. To assess the performance of our predictions, we used Mean Absolute Error (MAE) and Mean Absolute Percentage Error (MAPE) as evaluation metrics. These metrics help quantify the accuracy of our predictions by measuring the average magnitude of the errors in absolute terms and as a percentage of actual values, respectively. Table 2, and Table 3 are our experiment, measured by MAE and MAPE.

In addition to individual model predictions, we further improved the results of our experiment by leveraging the diverse strengths of multiple LMMs through LMM ensemble. In Table 3, we present the performance metrics of our LMM ensemble, demonstrating its effectiveness in eGFR prediction across different prompt responses, taking into account each LMM's contributions.

## V. DISCUSSION

From the results in Table 2., the most suitable prompts for predicting eGFR with each model are as follows:

**Gemini Flash**: Prompt 2 has the lowest MAE of 1.03 and MAPE of 3.59% on training, and MAE of 1.58 and MAPE of 5.57% on validation.

**Gemini Pro Vision**: Prompt 1 again shows the lowest MAE of 1.20 and MAPE of 5.47% on training, and MAE of 1.54 and MAPE of 4.98% on validation.

**GPT-4o**: Prompt 1 stands out with the lowest MAE of 2.04 and MAPE of 7.79% on training, and MAE of 2.06 and MAPE of 7.31% on validation.

**Claude 3 Opus**: Prompt 2 has the lowest MAE of 2.77 and MAPE of 12.40% on training, and MAE of 3.59 and MAPE of 16.87% on validation.

It is evident from the data that Gemini Flash, Gemini Pro Vision, and GPT-4o, when using their respective optimal prompts, outperform both Random Forest and 1D-CNN models. This observation underscores the superiority of LMMs in capturing the underlying patterns and relationships within the data, especially in predicting eGFR, compared to the traditional machine learning approaches. From Table 2 and Table 3 not only across different models but also within the LMMs ensemble, The LMMs ensemble with prompt 1 achieved results comparable to Random Forest. Although it is not significantly superior, it still represents a novel and pioneering attempt.

Drawing upon the importance of eGFR in assessing kidney function and monitoring CKD progression, this study showcases the potential of LMMs to enhance prediction accuracy. Accurate forecasting of eGFR trajectories is vital for timely intervention, tailored treatment plans, and improved patient outcomes. Additionally, understanding these trajectories facilitates effective kidney function management, allowing healthcare

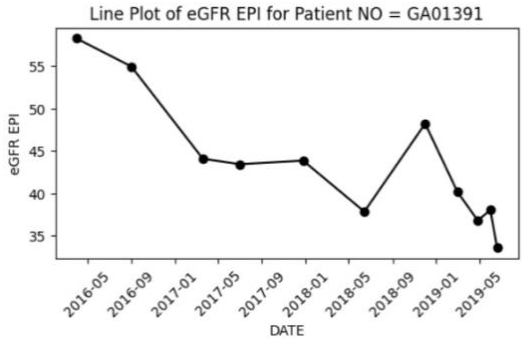

Fig. 2. Example of the line chart of eGFR trajectory of a patient.

TABLE III. RESULTS OF THE LMM ENSEMBLE METHOD.

| Prompt | Train | | Validation | |
| --- | --- | --- | --- | --- |
| | *MAE* | *MAPE(%)* | *MAE* | *MAPE(%)* |
| **prompt 1** | **2.21** | **8.97** | **2.81** | **10.26** |
| prompt 2 | 2.54 | 11.07 | 3.46 | 21.10 |
| prompt 3 | 4.10 | 25.12 | 4.12 | 19.55 |
| prompt 4 | 2.68 | 11.18 | 3.09 | 14.01 |

providers to arrange timely interventions and prepare for advanced CKD stages, such as ESRD.

By harnessing the capabilities of LMMs, this research lays the foundation for advancing medical forecasting using sophisticated computational models. Addressing the outlined limitations and refining the approach could lead to significant improvements in healthcare prediction tasks. Ultimately, these advancements have the potential to transform clinical decision-making processes, ultimately benefiting patients and enhancing the quality of care for those with CKD.

The limitations of this study include the limited size of the dataset, the use of manually created prompt templates, and reliance on a single type of graphical expression. By employing advanced prompt engineering techniques and auto-prompt generation, there is potential to extract more informative embeddings from LMMs. Additionally, different graphical representations might influence the LMMs' understanding of the problem, suggesting the need for a comprehensive experiment to identify the most effective visual formats. Future research endeavors could also encompass comparisons with LMMs in healthcare to further validate and refine the effectiveness of our approach. Moreover, exploring variations in prompts may enable LMMs to provide more nuanced and explainable predictions. Enhancing the explainability of LMM outputs can improve transparency and trust, making the models more valuable for practical applications.

## VI. Conclusion

In conclusion, our study demonstrates the superior predictive performance of LMMs compared to traditional predictive models. Notably, this research marks the first utilization of LMMs for eGFR prediction, showcasing their effectiveness in this domain. We found that prompt 1 of Gemini Pro Vision and prompt 2 of Gemini Flash consistently yielded promising results, indicating their efficacy in addressing eGFR-related queries. Through the utilization of LMMs, clinicians can gain deeper insights into the conditions of patients with CKD, ultimately improving their ability to provide optimal care.

### Acknowledgment

This study is mainly supported by the National Health Research Institutes (NHRI-EX113-11208PI).

APPENDIX: PROMPT TEMPLATES

In this study, we utilized four types of prompts: fill-in-the-blank, descriptive, open-ended, and role-playing. Each prompt template consists of a main body and variable names enclosed in curly brackets. For instance, the text "next_day_diff" within the curly brackets specifies the period until the patient's next visit. The text "eGFR" within double curly brackets marks the placeholder for LMMs to fill. The text "data_text" within curly brackets will be replaced by name-value pairs of other clinical and laboratory variables.

### A. Fill-in-the-blank Prompt

This prompt requires users to directly input specific information. The template used is: "Based on the provided plot, the x-axis represents the dates of each eGFR measurement, while the y-axis shows the eGFR values in units of mL/min/1.73m². This plot depicts the trajectory of a single patient's kidney function, as measured by the estimated Glomerular Filtration Rate (eGFR). Please fill in the blank: The most likely predicted value for the next {next_day_diff} days is {{eGFR}}mL/min/1.73m². The latest data for the patient: {data_text}."

### B. Descriptive Prompt

This prompt provides a detailed description of the plot and requests a prediction based on the data represented. The template is: "The x-axis represents the dates of each eGFR measurement, while the y-axis shows the eGFR values in units of mL/min/1.73m². This plot depicts the trajectory of a single patient's kidney function, as measured by the estimated Glomerular Filtration Rate (eGFR). Please provide the most likely predicted value for the next {next_day_diff} days as {{eGFR}}mL/min/1.73m². The latest data for the patient: {data_text}."

### C. Open-ended Prompt

This prompt allows users to analyze the data and identify trends in an open-ended manner without a structured response format. The template is: "The plot you see maps out the progression of kidney function for a single patient, using estimated Glomerular Filtration Rate (eGFR) values measured over various dates. Each point on the x-axis represents the date of measurement, and the corresponding y-axis value reflects the eGFR in mL/min/1.73m². Given the latest data provided: {data_text}, could you explore potential trends and predict how the patient's eGFR might evolve over the next {next_day_diff} days?"

### D. Role-playing Prompt

This prompt places the user in the role of a nephrologist tasked with making a prediction based on observed data and trends. The template is: "Imagine you are a nephrologist analyzing the patient's estimated Glomerular Filtration Rate trajectory. Based on your expertise, please predict the next {next_day_diff} days point's eGFR value in mL/min/1.73m² for this patient as {{eGFR}}mL/min/1.73m². Consider the trends and patterns observed in the plot, as well as any additional clinical information available. The latest data for the patient: {data_text}."